\documentclass{article}
\usepackage{spconf,amsmath,graphicx,hyperref}

\title{Flexi-LoRA with Input-Adaptive Ranks: \\ Efficient Finetuning for Speech and Reasoning Tasks}
\name{Zongqian Li, Yixuan Su, Han Zhou, Zihao Fu, Nigel Collier}
\address{University of Cambridge}
\usepackage{microtype}
\usepackage{inconsolata}
\usepackage{graphicx}
\usepackage{multirow}
\usepackage{booktabs}
\usepackage{amsmath}
\usepackage{algorithm}
\usepackage{algorithmic}
\usepackage{amsfonts}
\usepackage{subcaption}
\usepackage{xcolor}
\usepackage{listings}
\usepackage{enumitem}
\usepackage{soul}
\definecolor{myyellow}{rgb}{1, 0.972549, 0.584313}
\definecolor{mygreen}{rgb}{0.772549, 0.945098, 0.756862}
\definecolor{myblue}{rgb}{0.811764, 0.866666, 0.996078}
\definecolor{myred}{rgb}{0.984313, 0.749019, 0.737254}
\definecolor{darkgreen}{rgb}{0, 0.5001960, 0}
\definecolor{darkred}{rgb}{0.8, 0, 0}

\usepackage{colortbl}
\definecolor{marron}{HTML}{CC88B0}
\begin{document}
\topmargin=0mm
\ninept
\maketitle

\begin{abstract}
Parameter-efficient fine-tuning methods like Low-Rank Adaptation (LoRA) have become essential for deploying large language models, yet their static parameter allocation remains suboptimal for inputs of varying complexity. We present Flexi-LoRA, a novel framework that dynamically adjusts LoRA ranks based on input complexity during both training and inference. Through empirical analysis across question answering, mathematical reasoning, and speech tasks, we demonstrate that maintaining consistency between training and inference dynamics is important for effective adaptation, particularly for sequential reasoning tasks. Our findings reveal that input-dependent parameter allocation achieves higher performance with fewer parameters by optimally matching rank configurations to question complexity. Furthermore, task-specific dependency on rank dynamics varies, with mathematical reasoning tasks exhibiting higher dependency than QA tasks. Successful adaptation manifests not only in correctness but also in reasoning quality and instruction adherence. Flexi-LoRA consistently outperforms static LoRA while using fewer parameters, with performance gains more pronounced on tasks requiring strict reasoning chains. Our approach realizes key benefits of mixture-of-experts frameworks through a more streamlined implementation, reducing parameter redundancy while improving model capabilities. We provide comprehensive empirical studies across diverse tasks, establishing a basis for future work in input-adaptive and efficient fine-tuning approaches.
\footnote{\url{https://github.com/ZongqianLi/Flexi-LoRA}}
\end{abstract}


\section{Introduction}
\label{sec:intro}

As large language models grow in size, efficient fine-tuning methods like LoRA \cite{hu2022lora} have become essential for applications. However, their static parameter allocation remains suboptimal for questions of varying complexity, suggesting the need for input-adaptive approaches in parameter-efficient fine-tuning \cite{jiang2025finetuning}.

Through empirical analysis, we observe two key \textbf{phenomena} in LoRA-based fine-tuning. \textbf{First}, there exists a notable performance gap when using static ranks during inference for models trained with dynamic ranks at fine-grained level, particularly in their ability to follow instructions precisely (DyLoRA vs DyLoRA+, Table \ref{table-qa-cross-domain} and \ref{table-math}). \textbf{Second}, while model performance generally converges with increasing ranks, the optimal rank varies across different inputs: simple questions can be solved with small ranks, while complex problems benefit substantially from larger ranks (Rank 4 vs 8, Table \ref{table-qa-cross-domain} and \ref{table-math}). These observations indicate that a one-size-fits-all approach to rank selection is suboptimal, motivating the need for input-adaptive rank allocation.

\begin{figure}[t!]
    \centering
    \includegraphics[width=0.39\textwidth]{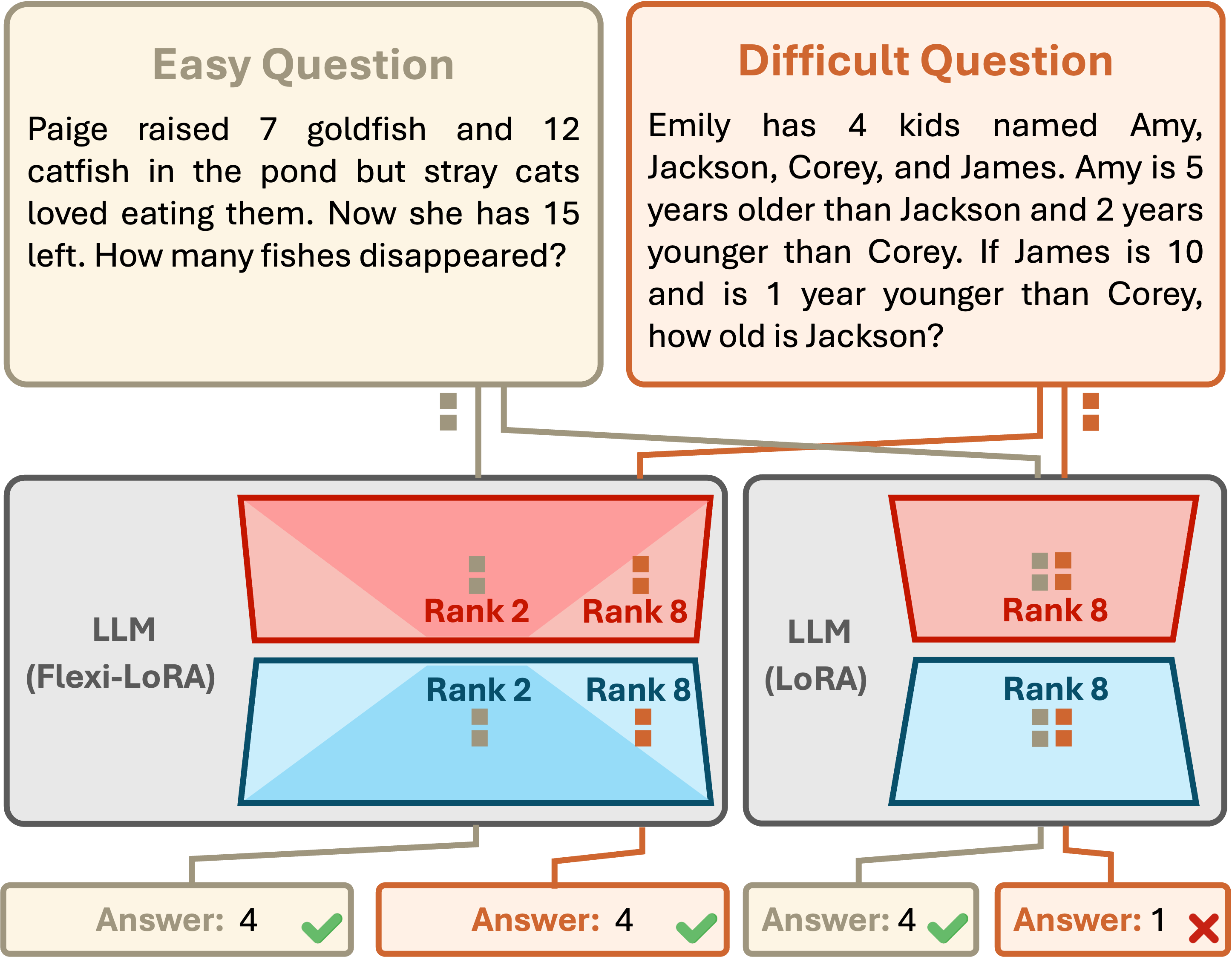}
    \caption{Flexi-LoRA's input-adaptive rank allocation versus static normal LoRA. Flexi-LoRA (left) dynamically assigns rank 2 (dark trapezoid) for simple problems and rank 8 (light trapezoid) for complex ones, successfully solving both. LoRA (right) uses fixed rank 8 (light trapezoid) regardless of complexity, failing on difficult problems. This demonstrates the necessity of input-adaptive parameter allocation for supporting varying question complexity.}
\label{cover_figure}
\vspace{-14pt}
\end{figure}

Inspired by these observations, we propose \textbf{Flexi-LoRA}, a finetuning framework that dynamically adjusts LoRA ranks based on input complexity. Our approach not only achieves higher performance to high-rank LoRA while using fewer parameters, but also successfully solves some complex problems that static LoRA fails to solve even with equivalent rank, as shown in Figure \ref{cover_figure}.

Our \textbf{contributions} are threefold:

\vspace{-6pt}

\begin{itemize}[left=0pt, itemsep=0pt, parsep=0pt]

\item \textbf{Novel Framework:} We introduce the first input-adaptive LoRA framework that maintains dynamic ranks during both training and inference, achieving higher performance with reduced parameter count compared to static LoRA.
\item \textbf{Insights:} We demonstrate that (1) maintaining consistency between training and inference dynamics is important for LoRA adaptation, particularly for sequential reasoning tasks; (2) input-dependent parameter allocation achieves high performance with fewer parameters by optimally matching rank configurations to question complexity; (3) task-specific dependency on rank dynamics varies, with mathematical reasoning tasks exhibiting higher dependency than QA tasks; (4) successful adaptation manifests not only in correctness but also in reasoning quality and instruction adherence; and (5) our approach realizes benefits of mixture-of-experts through a more streamlined implementation, reducing parameter redundancy while improving model capabilities.
\item \textbf{Comprehensive Analysis:} We provide comprehensive empirical studies across diverse tasks, establishing a basis for future work in input-adaptive and efficient finetuning approaches.

\end{itemize}

\section{Related Work}
\label{Related Work}

\textbf{LoRA with dynamic ranks.} Recent works have studied dynamic rank adaptation in LoRA, with differences shown in Table \ref{method_comparison}. AdaLoRA \cite{zhang2023adaptive} performs importance-based parameter reduction at training checkpoints to gradually reduce ranks to a fixed target. DyLoRA \cite{valipour-etal-2023-dylora} randomly samples ranks from a predefined range for each training batch, with all samples in the batch sharing the same rank. Both approaches, while improving rank flexibility, are limited by either steps-level reduction or random batch-level assignment, and neither supports dynamic rank selection at inference. On the other hand, Flexi-LoRA enables true sample-level rank selection by learning to map input complexity to appropriate ranks, maintaining this adaptive behavior during both training and inference.

\begin{table}[h!]
\centering
\setlength{\tabcolsep}{10pt}
\fontsize{7}{8}\selectfont
\begin{tabular}{ccccc}
\hline
\textbf{Method} & \textbf{Train} & \textbf{Level} & \textbf{Inference} \\
\hline
LoRA & Fixed & All & Fixed \\
AdaLoRA & Selective & Steps & Fixed \\
DyLoRA & Random & Batch & Fixed \\
\textbf{DyLoRA+ (Ours)} & Random & Batch & \textbf{Random} \\
\textbf{Flexi-LoRA (Ours)} & \textbf{Router} & \textbf{Sample} & \textbf{Router} \\
\hline
\end{tabular}
\caption{Comparison of rank adaptation strategies across different LoRA variants. "Train" indicates how ranks are determined during training, "Level" shows the level of rank assignment, and "Inference" specifies the rank selection method at test time. Only Flexi-LoRA maintains consistent router-based sample-level dynamic rank allocation across both training and inference stages, while existing methods use fixed ranks during inference regardless of training dynamics.}
\label{method_comparison}
\end{table}

\vspace{-12pt}

\section{Methods}
\label{Methods}

\begin{figure}[t!]
    \centering
    \includegraphics[width=0.35\textwidth]{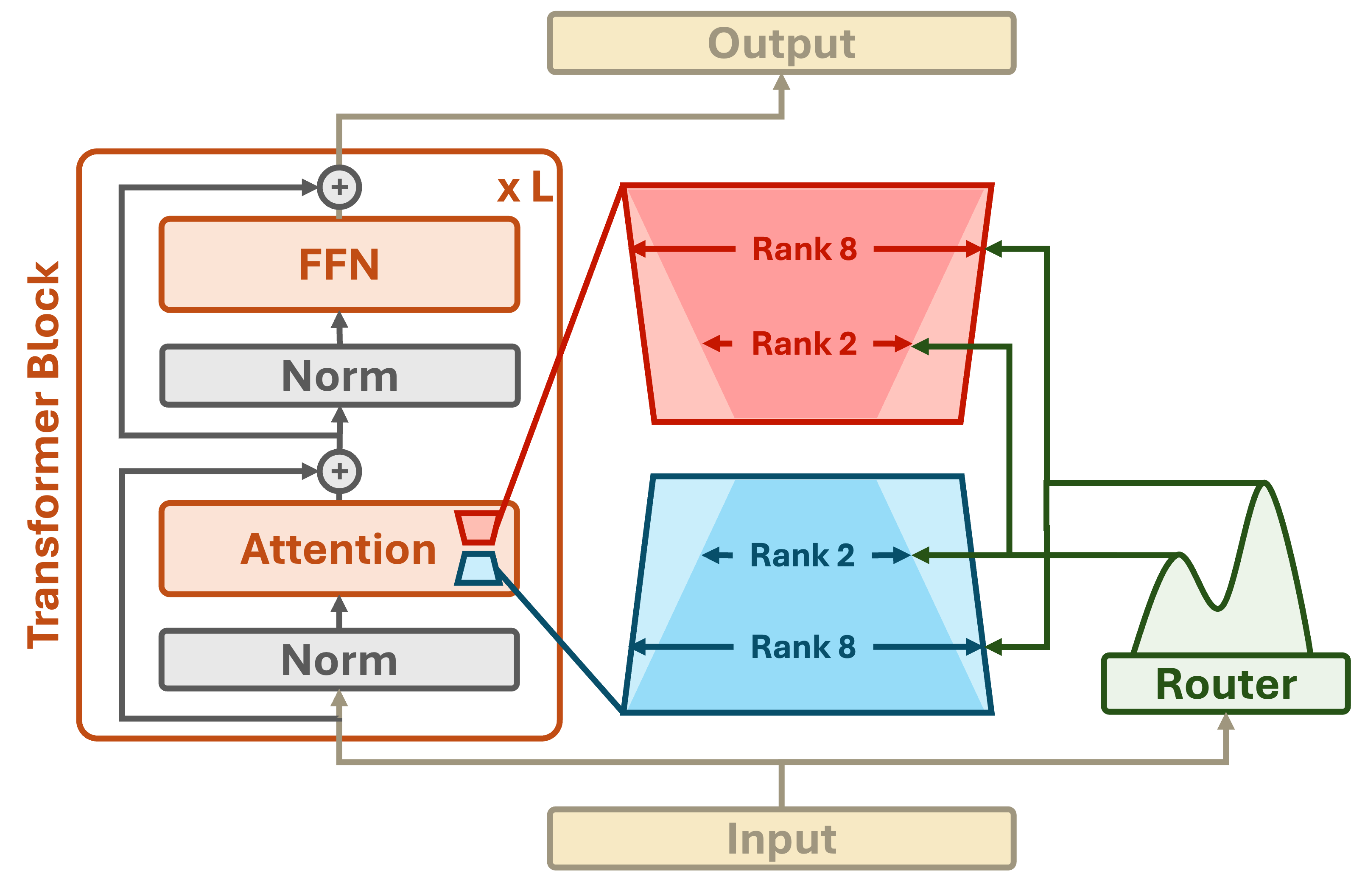}
    \caption{Flexi-LoRA framework with input-adaptive rank selection. Router analyzes input embeddings and outputs rank assignments (green arrows) for transformer layers. Red and blue trapezoids are LoRA's A and B matrices, with color darkness indicating rank magnitude (darker = rank 2, lighter = rank 8). The router enables dynamic rank allocation based on input complexity while maintaining efficient gradient flow through residual connections.}
    \label{method}
\end{figure}

Building upon previous work, we first introduce \textbf{DyLoRA+}, an improved variant of DyLoRA that maintains consistent rank dynamics by using random batch-level rank selection during both training and inference stages. While DyLoRA+ demonstrates improved performance over the original DyLoRA, its random rank allocation remains suboptimal as it fails to account for input-specific complexity differences. We therefore propose \textbf{Flexi-LoRA}, a framework that automatically adjusts the rank based on input complexity. Our method consists of two key components: a difficulty-aware router that maps inputs to appropriate rank assignments and a flexible-rank LoRA training framework that maintains consistent dynamic rank allocation during both training and inference, as shown in Figure \ref{method}.

\textbf{Router} focuses on learning an optimal mapping $R(h): \mathbb{R}^d \rightarrow {r_i}$ from input embeddings to rank assignments. Given an input sequence $x$ with mask $m$, we first compute its token embeddings $H \in \mathbb{R}^{n\times d}$ and obtain a pooled embedding $h = \sum_i(m_iH_i)/\sum_im_i$, where $m_i$ masks padding tokens. We categorize training samples into difficulty classes based on task-specific metrics: F1 scores for MRQA datasets and accuracy for mathematical reasoning tasks. The router is then optimized using a noise-added cross-entropy objective: $\mathcal{L}(\theta) = -\sum_i y_i\log(R(h_i + \epsilon))$, where $\epsilon \sim \mathcal{N}(0, \sigma^2)$ is Gaussian noise and $y_i$ denotes the ground-truth difficulty label. The training data is balanced between easy and hard samples to ensure uniform difficulty evaluation.

\textbf{Input-adaptive LoRA} freezes the base model parameters and optimize only the LoRA matrices. For input $x$, we first obtain its token embeddings $H^0$ and pooled embedding $h$ following the same procedure as router training. The router then predicts rank $r = R(h)$, which is applied consistently across all transformer layers. Within each batch, different samples can be assigned different ranks based on their predicted difficulty, enabling dynamic resource allocation. For each transformer layer $l$, we compute the LoRA update as $\Delta W_l = B_{l,r}A_{l,r}$, where $A_{l,r} \in \mathbb{R}^{r\times d}$ and $B_{l,r} \in \mathbb{R}^{d\times r}$ are dynamically reduced to the first $r$ rows/columns. The layer output is computed as $H^l = W_lH^{l-1} + \alpha_r(B_{l,r}A_{l,r}H^{l-1})$, where $\alpha_r$ is a rank-specific scaling variable and $H^{l-1}$ is the output from the previous layer. The model is trained to minimize the task-specific loss $\mathcal{L}_{\text{task}} = -\sum_i \log p(y_i|x_i)$, where $x_i$ is the input sequence and $y_i$ is the corresponding ground truth outputs. This design enables efficient batch processing while allowing for flexible input-dependent rank adaptation.

\begin{figure*}[t!]
    \centering
    \includegraphics[width=0.99\textwidth]{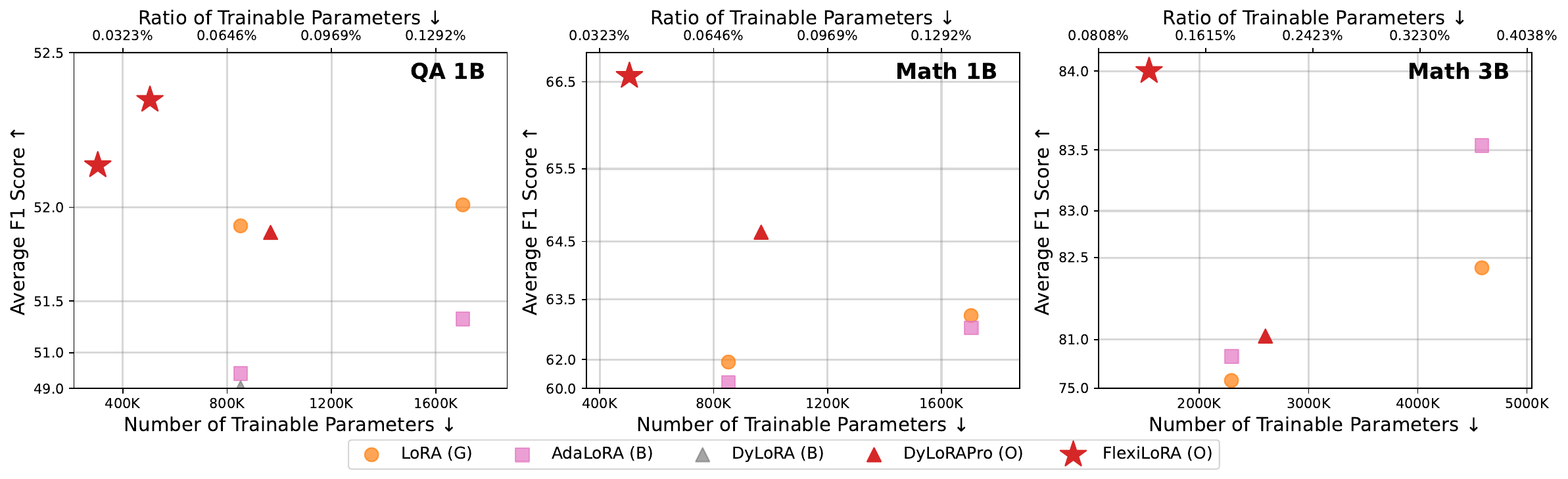}
    \caption{Performance-efficiency trade-off across different parameter-efficient fine-tuning methods. Our methods (O) achieve higher performance with fewer trainable parameters compared to both baseline methods (B) as well as gold standards (G). Results are shown for QA tasks and mathematical tasks using LLaMA-3.2-1B-Instruct and 3B.}
    \label{overall_results}
\end{figure*}

\section{Experimental Design}
\label{Experimental Design}

\textbf{Datasets.} We evaluate Flexi-LoRA on both QA and mathematical reasoning tasks. For QA tasks, we conduct training on datasets from the MRQA training set, which unifies QA samples from SQuAD \cite{rajpurkar-etal-2016-squad}, TriviaQA \cite{joshi-etal-2017-triviaqa}, NewsQA \cite{trischler-etal-2017-newsqa}, SearchQA \cite{dunn2017searchqanewqadataset}, HotpotQA \cite{yang-etal-2018-hotpotqa}, and NaturalQuestions \cite{kwiatkowski-etal-2019-natural}. Evaluation is performed on the MRQA test set consisting of BioASQ \cite{partalas2013results}, DROP \cite{dua-etal-2019-drop}, DuoRC \cite{saha-etal-2018-duorc}, RACE \cite{lai-etal-2017-race}, RelationExtraction \cite{levy2017zero}, and TextbookQA \cite{8100054}. For mathematical reasoning, we train on the GSM8K \cite{cobbe2021trainingverifierssolvemath} subset of the MetaMathQA dataset and evaluate on a diverse set of math benchmarks including GSM8K, SVAMP, MultiArith, and MAWPS. This design allows us to evaluate both in-distribution and out-of-distribution generalization capabilities of our method. For speech tasks, we use LibriSpeech.

\textbf{Evaluation Metrices.} We evaluate \textbf{QA} performance using \textbf{F1} and \textbf{Exact Match (EM)} scores. F1 computes the balanced average of precision and recall between prediction and ground truth, while EM measures exact string match. For \textbf{mathematical} reasoning tasks, we use \textbf{accuracy} for evaluation. For \textbf{speech}, we report Word Error Rate (\textbf{WER}), Character Error Rate (\textbf{CER}), and Accuracy (\textbf{ACC}).

\textbf{Gold Standard \& Baselines.} We compare against two \textbf{gold} standards: \textbf{full} model fine-tuning and standard \textbf{LoRA} with fixed rank. For \textbf{baselines}, we include \textbf{AdaLoRA}, which adapts ranks through importance-based parameter reduction while maintaining fixed inference ranks, and \textbf{DyLoRA}, which randomly samples ranks from a predefined range for each training batch but uses fixed ranks during inference. Our Flexi-LoRA differs by enabling input-adaptive rank selection during both training and inference.

\textbf{Models.} \textbf{LLaMA-3.2-1B-Instruct} \cite{grattafiori2024llama3herdmodels} is used as the base model for main results, and \textbf{LLaMA-3.2-3B-Instruct} is included to analyze model size in ablation studies. \textbf{Whisper} is used for speech tasks.

\section{Results}
\label{Results}

Figure \ref{overall_results} illustrates the performance-efficiency trade-offs across different parameter-efficient fine-tuning methods. Flexi-LoRA consistently achieves high performance while requiring fewer parameters than competing approaches: on QA tasks, Flexi-LoRA (2,8) achieves the highest average F1 (52.37\%) and EM (37.41\%) scores using only 29.59\% of LoRA-8's parameters, while on mathematical reasoning tasks, it attains 66.56\% accuracy (1B model) and 84.00\% accuracy (3B model), outperforming LoRA-8's 63.17\% and 82.37\% respectively with only 31.29\% and 33.40\% of its parameters. DyLoRA results are not visible in mathematical reasoning figures due to substantially decreased performance, highlighting the importance of training-inference consistency, while AdaLoRA shows competitive results on specific tasks but lacks consistent advantages across domains. From a Pareto optimality perspective, Flexi-LoRA dominates all baselines by offering better performance at lower parameter counts, positioning closest to full fine-tuning performance while maintaining parameter efficiency below 0.1\% of total model parameters.

\begin{table*}[t!]
    \centering
    \setlength{\tabcolsep}{12pt}
    \fontsize{7}{8}\selectfont
    \begin{tabular}{cccc|cccc|ccc}
        \hline
         & \multicolumn{3}{c|}{\textbf{Gold Standard}} & \multicolumn{4}{c|}{\textbf{Baseline}} & \multicolumn{3}{c}{\textbf{Ours}} \\
         & \textbf{Full} & \multicolumn{2}{c|}{\textbf{LoRA}} & \multicolumn{2}{c}{\textbf{AdaLoRA}} & \multicolumn{2}{c|}{\textbf{DyLoRA}} & \textbf{DyLoRA+} & \multicolumn{2}{c}{\textbf{Flexi-LoRA}} \\
        Rank & - & 4 & 8 & 4 & 8 & 4 & 8 & 1-8 & 1, 8 & 2, 8 \\
        \hline
        \# & 1.2B & 851K & 1703K & 851K & 1703K & 851K & 1703K & 966K & \textbf{304K} & \textbf{504K} \\
        \hline
        \multicolumn{11}{c}{\textbf{F1 Score}} \\
        \hline
        BioASQ & 69.81 & 64.85 & 66.22 & \cellcolor{marron!40}63.62 & \cellcolor{marron!20}65.12 & \cellcolor{marron!60}60.01 & \cellcolor{marron!60}52.40 & \cellcolor{teal!15}65.21 & \cellcolor{teal!45}65.75 & \cellcolor{teal!60}65.82 \\
        DROP & 47.15 & 37.88 & 36.03 & \cellcolor{marron!60}32.79 & \cellcolor{marron!40}34.35 & \cellcolor{teal!60}43.59 & \cellcolor{teal!45}38.26 & \cellcolor{marron!20}36.32 & \cellcolor{teal!15}37.27 & \cellcolor{teal!30}37.52 \\
        DuoRC & 45.21 & 43.85 & 43.72 & \cellcolor{teal!60}44.00 & \cellcolor{marron!20}42.49 & \cellcolor{marron!40}39.10 & \cellcolor{marron!60}36.89 & \cellcolor{teal!45}43.86 & \cellcolor{marron!20}42.92 & \cellcolor{teal!15}43.22 \\
        RACE & 41.49 & 38.64 & 37.49 & \cellcolor{marron!40}34.58 & \cellcolor{marron!20}35.81 & \cellcolor{teal!15}36.89 & \cellcolor{marron!60}33.24 & \cellcolor{teal!60}39.26 & \cellcolor{teal!30}38.83 & \cellcolor{teal!45}39.10 \\
        RE & 84.11 & 74.97 & 76.41 & \cellcolor{marron!40}74.51 & \cellcolor{marron!20}75.73 & \cellcolor{teal!60}81.02 & \cellcolor{teal!45}78.70 & \cellcolor{marron!60}75.47 & \cellcolor{teal!15}76.47 & \cellcolor{teal!30}76.83 \\
        TextbookQA & 49.56 & 51.35 & 52.20 & \cellcolor{teal!45}54.14 & \cellcolor{teal!60}54.66 & \cellcolor{marron!40}37.90 & \cellcolor{marron!60}32.94 & \cellcolor{marron!20}51.22 & \cellcolor{teal!15}51.74 & \cellcolor{teal!30}51.75 \\
        \hline
        Average & 56.22 & 51.92 & 52.01 & \cellcolor{marron!40}50.61 & \cellcolor{marron!20}51.36 & \cellcolor{marron!60}49.75 & \cellcolor{marron!60}45.40 & \cellcolor{teal!15}51.89 & \cellcolor{teal!45}52.16 & \cellcolor{teal!60}\textbf{52.37} \\
        \hline
        \multicolumn{11}{c}{\textbf{Exact Match}} \\
        \hline 
        BioASQ & 49.13 & 42.02 & 42.61 & \cellcolor{marron!40}40.29 & \cellcolor{marron!20}40.35 & \cellcolor{marron!60}34.24 & \cellcolor{marron!60}27.52 & \cellcolor{teal!60}41.88 & \cellcolor{teal!15}41.42 & \cellcolor{teal!45}41.48 \\
        DROP & 35.46 & 25.81 & 23.08 & \cellcolor{marron!60}20.15 & \cellcolor{marron!40}21.49 & \cellcolor{teal!60}30.07 & \cellcolor{marron!20}24.01 & \cellcolor{teal!15}23.61 & \cellcolor{teal!30}24.55 & \cellcolor{teal!45}25.01 \\
        DuoRC & 35.57 & 32.37 & 31.51 & \cellcolor{teal!60}32.37 & \cellcolor{teal!15}30.77 & \cellcolor{marron!40}27.71 & \cellcolor{marron!60}24.58 & \cellcolor{teal!45}32.31 & \cellcolor{marron!20}30.44 & \cellcolor{teal!30}30.71 \\
        RACE & 29.37 & 24.03 & 22.99 & \cellcolor{marron!40}20.62 & \cellcolor{marron!20}22.10 & \cellcolor{teal!15}21.81 & \cellcolor{marron!60}16.61 & \cellcolor{teal!60}24.92 & \cellcolor{teal!30}24.48 & \cellcolor{teal!45}24.77 \\
        RE & 72.01 & 57.73 & 60.27 & \cellcolor{marron!40}57.25 & \cellcolor{marron!20}58.44 & \cellcolor{teal!60}69.13 & \cellcolor{teal!45}66.31 & \cellcolor{marron!60}59.15 & \cellcolor{teal!15}59.90 & \cellcolor{teal!30}60.44 \\
        TextbookQA & 40.98 & 41.91 & 42.38 & \cellcolor{teal!45}44.97 & \cellcolor{teal!60}45.10 & \cellcolor{marron!40}27.01 & \cellcolor{marron!60}21.29 & \cellcolor{marron!20}41.91 & \cellcolor{teal!15}42.04 & \cellcolor{teal!30}42.04 \\
        \hline
        Average & 43.75 & 37.31 & 37.14 & \cellcolor{marron!40}35.94 & \cellcolor{marron!20}36.38 & \cellcolor{marron!60}34.99 & \cellcolor{marron!60}30.05 & \cellcolor{teal!45}37.30 & \cellcolor{teal!15}37.14 & \cellcolor{teal!60}\textbf{37.41} \\
        \hline
    \end{tabular}
    \caption{Performance comparison on out-of-domain \textbf{QA} tasks from the MRQA benchmark using LLaMA-3.2-1B-Instruct. F1 and Exact Match (EM) scores are reported, comparing our proposed methods (Flexi-LoRA and DyLoRA+) against gold standards and baselines. Flexi-LoRA (2,8) achieves the best average performance on both metrics while using fewer parameters than standard approaches. The "\#" row indicates the number of trainable parameters. Green (teal) and red (maroon) cell coloring is higher and lower scores respectively, with deeper colors indicating larger performance differences.}
    \label{table-qa-cross-domain}
\end{table*}

\begin{table*}[t!]
    \centering
    \setlength{\tabcolsep}{13pt}
    \fontsize{7}{8}\selectfont
    \begin{tabular}{cccc|cccc|cc}
        \hline
         & \multicolumn{3}{c|}{\textbf{Gold Standard}} & \multicolumn{4}{c|}{\textbf{Baseline}} & \multicolumn{2}{c}{\textbf{Ours}} \\
         & \textbf{Full} & \multicolumn{2}{c|}{\textbf{LoRA}} & \multicolumn{2}{c}{\textbf{AdaLoRA}} & \multicolumn{2}{c|}{\textbf{DyLoRA}} & \textbf{DyLoRA+} & \multicolumn{1}{c}{\textbf{Flexi-LoRA}} \\
        Rank & - & 4 & 8 & 4 & 8 & 4 & 8 & 1-8 & 2, 8 \\
        \hline
        \multicolumn{10}{c}{\textbf{LLaMA-3.2-1B-Instruct}} \\
        \hline
        \# & 1.2B & 851K & 1703K & 851K & 1703K & 851K & 1703K & 953K & \textbf{533K} \\
        \hline
        GSM8K & 57.31 & 42.15 & 41.31 & \cellcolor{teal!60}45.71 & \cellcolor{teal!20}42.22 & \cellcolor{marron!60}19.78 & \cellcolor{marron!40}21.00 & \cellcolor{marron!20}41.77 & \cellcolor{teal!40}42.30 \\
        SVAMP & 57.29 & 52.87 & 51.18 & \cellcolor{teal!40}53.37 & \cellcolor{marron!20}50.52 & \cellcolor{marron!40}20.08 & \cellcolor{marron!60}19.57 & \cellcolor{teal!60}56.03 & \cellcolor{teal!20}52.02 \\
        MultiArith & 93.88 & 82.77 & 85.00 & \cellcolor{marron!20}78.88 & \cellcolor{teal!20}84.44 & \cellcolor{marron!40}43.88 & \cellcolor{marron!60}42.77 & \cellcolor{teal!40}85.55 & \cellcolor{teal!60}92.22 \\
        MAWPS & 80.00 & 69.85 & 75.21 & \cellcolor{marron!20}65.63 & \cellcolor{teal!20}74.36 & \cellcolor{marron!40}22.25 & \cellcolor{marron!60}17.46 & \cellcolor{teal!40}75.21 & \cellcolor{teal!60}79.71 \\
        \hline
        Average & 72.12 & 61.91 & 63.17 & \cellcolor{marron!20}60.90 & \cellcolor{teal!20}62.89 & \cellcolor{marron!40}26.50 & \cellcolor{marron!60}25.20 & \cellcolor{teal!40}64.64 & \cellcolor{teal!60}\textbf{66.56} \\
        \hline
        \multicolumn{10}{c}{\textbf{LLaMA-3.2-3B-Instruct}} \\
        \hline
        \# & 1.2B & 2.29M & 4.58M & 2.29M & 4.58M & 2.29M & 4.58M & 2.60M & \textbf{1.53M} \\
        \hline
        GSM8K & 75.20 & 65.95 & 69.37 & \cellcolor{teal!40}71.49 & \cellcolor{teal!60}72.32 & \cellcolor{marron!60}60.04 & \cellcolor{marron!40}60.87 & \cellcolor{teal!20}70.43 & \cellcolor{marron!20}69.90 \\
        SVAMP & 78.30 & 66.29 & 74.47 & \cellcolor{teal!20}75.91 & \cellcolor{teal!60}77.80 & \cellcolor{marron!40}67.48 & \cellcolor{marron!60}64.87 & \cellcolor{marron!20}74.71 & \cellcolor{teal!40}77.09 \\
        MultiArith & 100 & 97.77 & 99.44 & \cellcolor{marron!20}90.55 & \cellcolor{teal!40}95.55 & \cellcolor{marron!60}75.00 & \cellcolor{marron!40}82.22 & \cellcolor{teal!20}92.77 & \cellcolor{teal!60}100 \\
        MAWPS & 89.85 & 84.50 & 86.19 & \cellcolor{marron!20}83.66 & \cellcolor{teal!40}88.45 & \cellcolor{marron!40}69.57 & \cellcolor{marron!60}65.63 & \cellcolor{teal!20}86.47 & \cellcolor{teal!60}89.01 \\
        \hline
        Average & 85.84 & 78.63 & 82.37 & \cellcolor{marron!20}80.40 & \cellcolor{teal!40}83.53 & \cellcolor{marron!60}68.02 & \cellcolor{marron!40}68.40 & \cellcolor{teal!20}81.10 & \cellcolor{teal!60}\textbf{84.00} \\
        \hline
    \end{tabular}
    \caption{Performance comparison on \textbf{mathematical tasks} using LLaMA-3.2-1B-Instruct and LLaMA-3.2-3B-Instruct models. Accuracy scores (\%) are reported across four benchmark datasets (GSM8K as in-domain and the others as out-of-domain), comparing our proposed methods against gold standards and baselines. Flexi-LoRA achieves the best average performance on both model sizes while using fewer parameters than standard approaches.}
    \label{table-math}
    \vspace{-6pt}
\end{table*}

\begin{table}[h!]
\centering
\setlength{\tabcolsep}{12.5pt}
\fontsize{7}{8}\selectfont
\begin{tabular}{ccccc}
\hline
& \multicolumn{3}{c|}{\textbf{Gold Standard}} & \multicolumn{1}{c}{\textbf{Ours}} \\
\textbf{} & \textbf{Full} & \multicolumn{2}{c|}{\textbf{LoRA}} & \textbf{FlexiLoRA} \\
Rank & - & 4 & \multicolumn{1}{c|}{8} & 2, 8 \\
\#\%  $\downarrow$ & 100 & 0.29 & \multicolumn{1}{c|}{0.58} & \textbf{0.15} \\
\hline
\multicolumn{5}{c}{\textbf{Clear}} \\
\hline
WER $\downarrow$ & 13.45 & \cellcolor{marron!20}17.82 & \multicolumn{1}{c|}{\cellcolor{marron!60}17.85} & \cellcolor{teal!60}\textbf{14.33} \\
CER $\downarrow$ & 4.75 & \cellcolor{marron!20}5.28 & \multicolumn{1}{c|}{\cellcolor{marron!60}5.30} & \cellcolor{teal!60}\textbf{4.62} \\
ACC $\uparrow$ & 86.55 & \cellcolor{marron!20}82.18 & \multicolumn{1}{c|}{\cellcolor{marron!60}82.15} & \cellcolor{teal!60}\textbf{85.67} \\
\hline
\multicolumn{5}{c}{\textbf{Other}} \\
\hline
WER $\downarrow$ & 23.86 & \cellcolor{marron!60}27.33 & \multicolumn{1}{c|}{\cellcolor{marron!20}27.23} & \cellcolor{teal!60}\textbf{23.85} \\
CER $\downarrow$ & 11.76 & \cellcolor{marron!60}11.35 & \multicolumn{1}{c|}{\cellcolor{marron!20}11.29} & \cellcolor{teal!60}\textbf{10.17} \\
ACC $\uparrow$ & 76.14 & \cellcolor{marron!60}72.67 & \multicolumn{1}{c|}{\cellcolor{marron!20}72.77} & \cellcolor{teal!60}\textbf{76.15} \\
\hline
\multicolumn{5}{c}{\textbf{All}} \\
\hline
WER $\downarrow$ & 18.66 & \cellcolor{marron!60}22.58 & \multicolumn{1}{c|}{\cellcolor{marron!20}22.54} & \cellcolor{teal!60}\textbf{19.09} \\
CER $\downarrow$ & 8.26 & \cellcolor{marron!60}8.32 & \multicolumn{1}{c|}{\cellcolor{marron!20}8.30} & \cellcolor{teal!60}\textbf{7.40} \\
ACC $\uparrow$ & 81.35 & \cellcolor{marron!60}77.43 & \multicolumn{1}{c|}{\cellcolor{marron!20}77.46} & \cellcolor{teal!60}\textbf{80.91} \\
\hline
\end{tabular}
\caption{Performance across \textbf{speech tasks} on LibriSpeech.}
\label{table-speech}
\end{table}

Table \ref{table-qa-cross-domain} presents the performance of different parameter-efficient fine-tuning methods on six out-of-domain QA datasets from the MRQA benchmark. We analyze these results from multiple perspectives:
(1) \textbf{Stability Across Metrics}: Flexi-LoRA demonstrates consistency by achieving the best average performance on both F1 and EM metrics simultaneously, unlike other methods that typically excel in one metric. This dual-metric advantage indicates that Flexi-LoRA produces outputs that are both semantically close to ground truth (high F1) and syntactically precise (high EM).
(2) \textbf{Dataset-Specific Analysis}: Flexi-LoRA performs particularly well on domain-specific and knowledge-based tasks, while showing moderate improvements on information extraction tasks. The adaptive rank allocation proves beneficial for datasets requiring diverse reasoning capabilities, suggesting Flexi-LoRA's ability to assign appropriate computational resources based on input complexity.
(3) \textbf{Comparison with Other Dynamic Approaches}: Flexi-LoRA consistently outperforms DyLoRA+, despite both using dynamic ranks during inference, demonstrating the importance of learned input-adaptive rank assignment versus the random assignment in DyLoRA+. DyLoRA shows high variance across datasets (from 81.02\% F1 on RE to 32.94\% F1 on TextbookQA), highlighting variability when training and inference dynamics are inconsistent.
(4) \textbf{Cross-Domain Generalization}: On out-of-domain test datasets, baseline methods exhibit inconsistent performance, excelling on specific domains while decreasing on others. On the other hand, Flexi-LoRA maintains strong performance across all test datasets, suggesting its input-adaptive parameter allocation learns more generalized knowledge than static approaches that could overfit to training domain characteristics.
(5) \textbf{Key Insights}: These results demonstrate that input-adaptive parameter allocation provides dual benefits in QA tasks: improved performance through parameter allocation and improved parameter efficiency through optimization of rank selection. The consistent improvement across diverse datasets suggests that question complexity varies substantially even within the same task category, validating our input-adaptive approach.

Tables \ref{table-math} and \ref{table-speech} present the performance of different fine-tuning methods on mathematical reasoning tasks across two model sizes and on speech tasks. We analyze these results for task-specific characteristics:
(1) \textbf{Model Size Influence}: Increasing model size from 1B to 3B improves performance across all methods, with Flexi-LoRA maintaining its advantage. Notably, the absolute performance gap between Flexi-LoRA and full fine-tuning decreases from 5.56\% to 1.84\%, suggesting that input-adaptive approaches become even more effective with larger models.
(2) \textbf{In-Domain vs. Out-of-Domain}: Flexi-LoRA shows strong generalization from GSM8K (in-domain) to out-of-domain datasets. On the 1B model, Flexi-LoRA achieves an average accuracy of 74.65\% on out-of-domain tasks (SVAMP, MultiArith, MAWPS), compared to 42.30\% on in-domain GSM8K, demonstrating cross-domain robustness. This is consistent across both model sizes.
(3) \textbf{Dataset Complexity}: Flexi-LoRA performs well on both elementary arithmetic (MultiArith) and complex multi-step reasoning (GSM8K), indicating its ability to support varying levels of mathematical complexity. The high performance on MultiArith (92.22\%) approaches full fine-tuning (93.88\%), showcasing the method's ceiling capability on reasoning tasks.
(4) \textbf{DyLoRA Performance Decrease}: DyLoRA exhibits performance decrease on mathematical tasks (average 26.50\% on 1B model). The 40.06\% performance gap between DyLoRA and Flexi-LoRA on math tasks highlights the influence of training-inference inconsistency on sequential reasoning tasks. This decrease is substantially more pronounced than in QA tasks, suggesting that mathematical reasoning is particularly dependent on dynamic rank consistency.
(5) \textbf{Key Insights}: These results demonstrate that maintaining consistent training-inference dynamics is important for mathematical reasoning tasks. The substantial performance improvements (Flexi-LoRA outperforms LoRA-8 by 3.39\% on 1B) illustrate that input-adaptive parameter allocation provides greater benefits for problems with higher complexity variance and stricter evaluation criteria, compared to the modest gains observed on QA tasks.

\section{Conclusions}
\label{Conclusions}

This paper introduces Flexi-LoRA, an input-adaptive framework that dynamically adjusts LoRA ranks based on question complexity. We demonstrate that maintaining consistent rank dynamics between training and inference is important for finetuning models, particularly for sequential reasoning tasks. Flexi-LoRA outperforms static LoRA while using fewer parameters (29.6\% for QA and 31.3\% for math reasoning), with performance gains more pronounced on mathematical tasks requiring reasoning chains. These results confirm that input-dependent parameter allocation enables efficient capacity allocation while reducing parameter redundancy, achieving benefits similar to mixture-of-experts frameworks through a more streamlined method.

Future work could have several directions: (1) layer-specific dynamic ranks to optimize parameter utilization at a finer level; (2) router frameworks learning detailed aspects of input complexity; (3) combination with other parameter-efficient methods such as sparse fine-tuning. Given its efficiency and high performance, Flexi-LoRA has the potential to be used in different areas and applications.

\vfill\pagebreak

\bibliographystyle{IEEEtran}
\bibliography{strings,refs}

\end{document}